\documentclass[times,review,10pt]{elsarticle}

\usepackage{amssymb}
\usepackage{amsmath}
\usepackage{url}
\usepackage{caption}
\usepackage{pifont}
\usepackage{makecell}
\usepackage{booktabs}
\usepackage{multirow}
\usepackage[]{hyperref}
\usepackage{xcolor}

\makeatletter
\newcommand{\eadabbr}[2]{%
  \begingroup
  \@eadauthor={#2}%
  \ead{#1}%
  \endgroup
}
\makeatother

\begin{document}

\begin{frontmatter}



\title{DBCF: Dual-Branch Complementary Fusion of Foundation Models for Generalized Deepfake Detection}

\author[inst1,inst2]{Fengming Gu}
\eadabbr{gufengming18@mails.ucas.ac.cn}{F. Gu}

\author[inst2,inst3]{Mingjie He}
\eadabbr{hemingjie@ict.ac.cn}{M. He}

\author[inst4]{Zonghui Guo}
\eadabbr{guozonghui@ouc.edu.cn}{Z. Guo}

\author[inst2,inst3]{Jie Zhang\corref{cor1}}
\eadabbr{zhangjie@ict.ac.cn}{J. Zhang}

\author[inst2,inst3]{Shiguang Shan}
\eadabbr{sgshan@ict.ac.cn}{S. Shan}

\cortext[cor1]{Corresponding author.}

\affiliation[inst1]{organization={School of Advanced Interdisciplinary Sciences, University of Chinese Academy of Sciences},
            city={Beijing}, 
            postcode={100049}, 
            country={China}}
            
\affiliation[inst2]{organization={State Key Laboratory of AI Safety, Institute of Computing Technology,\\ Chinese Academy of Sciences},
            city={Beijing}, 
            postcode={100190}, 
            country={China}}

\affiliation[inst3]{organization={University of Chinese Academy of Sciences},
            city={Beijing}, 
            postcode={100049}, 
            country={China}}

\affiliation[inst4]{organization={Faculty of Information Science and Engineering, Ocean University of China},
            city={Qingdao}, 
            postcode={266404}, 
            country={China}}

\begin{abstract}
As image generation and editing technologies have progressed substantially, facial forgeries pose significant challenges to privacy and public safety. Due to limited ability to capture forgery cues, existing small-scale forgery detection models often struggle to generalize across various domains and unseen manipulations. To address this limitation, researchers have turned to large-scale foundation models, which can provide richer representations and better generalization. Nevertheless, relying on a single foundation model alone remains insufficient for effective forgery detection. While models like CLIP offer robust global semantic cues, they lack the capacity to capture detailed local facial features. In contrast, DINO excels at capturing local structural features of faces, but provides weaker global semantic context. To fully utilize the synergies among multiple foundation models, we propose a hierarchical multi-granular framework that integrates complementary pretrained representations. Specifically, a Global Context Branch (GCB) based on CLIP captures holistic semantic cues, while a Fine-grained Cue Branch (FCB) built on DINOv3 captures localized structural irregularities. In addition, we design a feature fusion module that enables parameter-efficient adaptation of the frozen foundation backbones by adaptively extracting and integrating complementary features from the two models. By jointly leveraging global context and fine-grained cues, our method learns more comprehensive forgery representations and achieves strong cross-manipulation performance. Extensive experiments on multiple benchmarks demonstrate the benefit of the proposed design, particularly under cross-dataset and cross-manipulation settings.
\end{abstract}




\end{frontmatter}



\section{Introduction}
\label{sec1}

The rapid advancement of artificial image generation technologies, particularly face manipulation and synthesis methods based on generative adversarial networks (GANs) and diffusion models, has profoundly reshaped the landscape of digital media creation. While these techniques have enabled numerous beneficial applications, they have also drastically lowered the barrier for producing highly realistic facial forgeries. Such forgeries pose severe 
threats to personal privacy, social trust, and even public security, as they can be exploited for identity fraud, political misinformation, and malicious impersonation. Consequently, facial deepfake detection has emerged as an
essential research task aimed at safeguarding the integrity and authenticity of visual content.

However, building robust and generalizable deepfake detectors remains highly challenging. The core difficulty lies in the diversity and rapid evolution of forgery techniques, which continuously introduce novel manipulation patterns. In response to such challenges, several prior methods\cite{li2020face,shiohara2022detecting,yan2024transcending} generate diverse manipulated samples that approximate potential unseen forgery patterns. Besides, some other methods\cite{yan2023ucf,zhuang2022uia} employ disentangled representation learning which can suppress irrelevant factors such as identity, illumination, or background that may act as shortcuts rather than genuine forgery cues. To some extent, these methods can prevent the models from learning only limited, dataset-specific cues from known manipulations. However, their performance is still largely constrained by the scale of the training data and the distribution of the manipulation samples.

Recently, a promising complementary direction has emerged by leveraging large-scale pretrained vision models for forgery detection. These foundation models encode rich visual priors from massive and diverse datasets. Among them, CLIP has attracted considerable attention because its contrastive image-text pretraining provides robust global semantic representations. Recent studies~\cite{shao2025deepfake} demonstrate that using CLIP as the backbone for deepfake detection can enhance generalization across unseen manipulations by exploiting holistic global context.

Despite their strong generalization ability, directly applying pretrained vision models such as CLIP to facial forgery detection remains nontrivial. Owing to their representation characteristics, CLIP-based detectors often exhibit limited sensitivity to subtle, low-level forgery cues. This limitation primarily stems from the contrastive pretraining objective, which emphasizes global semantic alignment over localized visual inconsistencies. As noted in prior work~\cite{shao2025deepfake}, this strong global context bias, while beneficial for generalization, can be ineffective for manipulations that introduce only minor local artifacts (e.g., NeuralTexture~\cite{thies2019deferred}) without altering overall semantics, resulting in suboptimal detection of fine-grained local cues.

To overcome this limitation, we seek to integrate complementary pretrained representations in a principled manner, thereby balancing holistic semantic understanding with sensitivity to subtle manipulation cues. We find that the self-supervised backbone DINOv3~\cite{simeoni2025dinov3} is particularly well suited for modeling local structural patterns and fine-grained visual details. It has potential to complement CLIP, which primarily encodes global semantic context.

To this end, we propose a dual-branch multi-granular feature fusion framework designed to integrate two pretrained vision models, CLIP and DINOv3, while preserving their pretrained representations and enabling complementary feature fusion. Crucially, these two models are pretrained under fundamentally different learning paradigms, and their feature spaces are therefore not inherently aligned. In the absence of explicit alignment or regulation, naively fusing these representations or jointly finetuning them can erode the structured feature patterns induced by pretraining and compromise the effective use of pretrained priors. Specifically, DBCF keeps the pretrained backbones frozen and learns complementary interactions at the representation level, which helps preserve pretrained priors while reducing the risk of overfitting to dataset-specific artifacts. This design allows the model to exploit both global semantic consistency and fine-grained manipulation traces without directly collapsing the heterogeneous feature spaces of CLIP and DINOv3.

Extensive experiments conducted on several widely used deepfake detection benchmarks demonstrate the effectiveness of our approach. In particular, it achieves strong performance in both cross-dataset and cross-manipulation evaluations when compared with prior approaches that use only frozen or fully fine-tuned pretrained backbones These results support our hypothesis that combining complementary pretrained backbones provides a principled and robust pathway toward generalizable deepfake detection.

In summary, the main contributions of this work are threefold:

1. We introduce a dual-branch, multi-granular framework that leverages the complementary strengths of CLIP and DINOv3. It incorporates a spatially aligned fusion mechanism to integrate global and local representations while preserving pretrained priors.

2.	We provide an in-depth analysis of the inherent limitations in directly adapting pretrained visual models for deepfake detection, revealing a fundamental tension between preserving robust global semantic representations and enhancing sensitivity to fine-grained manipulation cues.

3.	We conduct extensive evaluations on multiple datasets and forgery types, demonstrating that our approach achieves superior robustness and generalization performance compared with existing state-of-the-art methods.

\section{Related Works}

\textbf{Face Forgery Generation.} 
Deepfake refers to AI-generated forgeries produced by deep generative models, capable of synthesizing, modifying, or replacing human-centric visual or auditory content. Existing deepfake generation methods can be broadly categorized into three main types: face swapping, face reenactment, and entire face synthesis. Among them, face swapping~\cite{chen2020simswap} techniques are widely used to transfer the source identity onto a target individual’s appearance, motion, and scene context ~\cite{liu2023deepfacelab}, producing videos where the target person convincingly appears as the source.
In contrast, face reenactment focuses on transferring facial motion while preserving the target identity, allowing the target face to mimic expression, pose, or lip movements driven by various modalities such as images ~\cite{siarohin2019first} and audio ~\cite{prajwal2020lip}.
These approaches typically disentangle and manipulate identity-related and motion-related representations. Beyond manipulation-based pipelines, entire face synthesis methods generate photorealistic facial images from learned generative distributions. Representative GAN-based models, such as the StyleGAN family~\cite{karras2020analyzing,karras2021alias}, enable high-fidelity and controllable face generation, while transformer-assisted models like VQGAN~\cite{esser2021taming} synthesize high-resolution images via discrete tokens. Diffusion-based methods, benefiting from advanced diffusion architectures, further provide more realistic and controllable synthesis, as exemplified by Stable Diffusion~\cite{rombach2022high}.
For clarity, entire face synthesis aligns more closely with generic image synthesis detection and thus falls outside the scope of the manipulation-based face forgery task addressed in this work.

\textbf{Face Forgery Detection.}
Due to the rapid emergence of numerous forgery techniques, a large body of research has focused on improving the generalization capability of deepfake detectors against unseen forgery methods. Methods based on data augmentation enhance generalization by synthesizing diverse and transferable forgery patterns. Face X-ray~\cite{li2020face} and SBI~\cite{shiohara2022detecting} augment samples at the image level, while LSDA~\cite{yan2024transcending} enriches forgery variations in the feature space. These techniques have been shown to be effective by exposing detectors to a broader distribution of manipulation artifacts. Another line of work improves domain generalization through feature disentanglement and real-face modeling. Disentanglement~\cite{yan2023ucf} eliminates the influence of irrelevant features, while real-face representation learning~\cite{shi2025real} enhances cross-manipulation robustness.
Additionally, methods like facial landmarks~\cite{gao2025leveraging} and mask-guided supervision~\cite{li2025deepfake} guide the model to focus on relevant features, while multi-definition cross-domain training~\cite{zhao2025multi} enhances robustness to low-quality or previously unseen deepfakes.

Complementary feature modeling has also been investigated in recent hybrid
frameworks. For example, LGDF-Net~\cite{long2025lgdf} introduces local and
global branches to capture localized artifacts and global facial
texture/context through multi-scale and multi-level fusion, while Ding et
al.~\cite{ding2025face} jointly exploit RGB and noise-map representations
across multiple scales. These methods demonstrate the effectiveness of
incorporating complementary cues through task-oriented architectural designs.
Beyond architectural specialization for complementary cue extraction, the
integration of complementary transferable representations from pretrained
models represents another promising direction for generalized deepfake
detection.

More recently, researchers have begun to explore the potential of large pretrained models for generalizable deepfake detection. For instance, CLIPping~\cite{khan2024clipping} and UniFD~\cite{ojha2023towards} demonstrate that it is feasible to equip CLIP with universal deepfake detection capabilities through tailored fine-tuning strategies. In addition, methods such as Ffaa~\cite{huang2024ffaa} and $\mathcal{X}^{2}$-DFD~\cite{chenmathcal} further investigate the effectiveness of vision--language models (VLMs) in this task.
Overall, these studies demonstrate the strong transferability of large-scale pretrained representations for generalized deepfake detection, providing a foundation for exploring more effective ways to exploit pretrained visual priors in this task.

\textbf{Vision Foundation models.}
Vision Foundation models (VFMs) have become a cornerstone of modern computer vision. After training on large-scale and diverse datasets, VFMs have acquired rich and transferable visual knowledge, yielding strong performance across a wide range of downstream tasks. Benefiting from the global contextual attention mechanism, Vision Transformer (ViT)~\cite{dosovitskiy2020vit} is adopted as the base architecture of most VFMs. 
The success of ViT-based VFMs can be attributed to two large-scale pretraining paradigms, i.e. Cross-Modal Supervised Learning and Visual Self-Supervised Learning. These two paradigms focus on distinct types of feature extraction. The first paradigm utilizes cross-modal (vision-text) contrastive learning for supervision. This approach, exemplified by the CLIP~\cite{radford2021learning} family  effectively extends the definition of supervised learning beyond hard classification labels. This cross-modal alignment grants these models strong global semantic consistency and impressive zero-shot transferability. 
Another critical paradigm focuses on Visual Self-Supervised Learning (SSL), where models generate their own supervisory signals exclusively from image data through pretext tasks. These methods can be broadly categorized into: those based on instance discrimination via contrastive techniques (e.g., MoCo~\cite{he2020momentum}), those utilizing masked image modeling (e.g., MAE~\cite{he2022masked}), and those using non-contrastive clustering or distillation techniques (e.g., DINO~\cite{oquab2024dinov2}). These VFMs are trained to understand the image's intrinsic structure without relying on external labels. Thanks to the novel Gram-Anchoring regularization, the latest DINOv3~\cite{simeoni2025dinov3} exhibits exceptional fidelity in modeling local structural patterns and fine-grained visual cues. This strong emphasis on internal structural consistency enables SSL-based visual models to adapt effectively to tasks such as object detection and semantic segmentation.
While cross-modal supervised and self-supervised VFMs excel in different aspects of visual understanding, their complementary characteristics motivate a dual-branch approach that integrates global semantics and fine-grained local cues for improved generalization in forgery detection.

\section{Method}

\subsection{Overview}

\begin{figure}[t]
\centering
\includegraphics[width=1.0\linewidth, trim=80 60 120 30, clip]{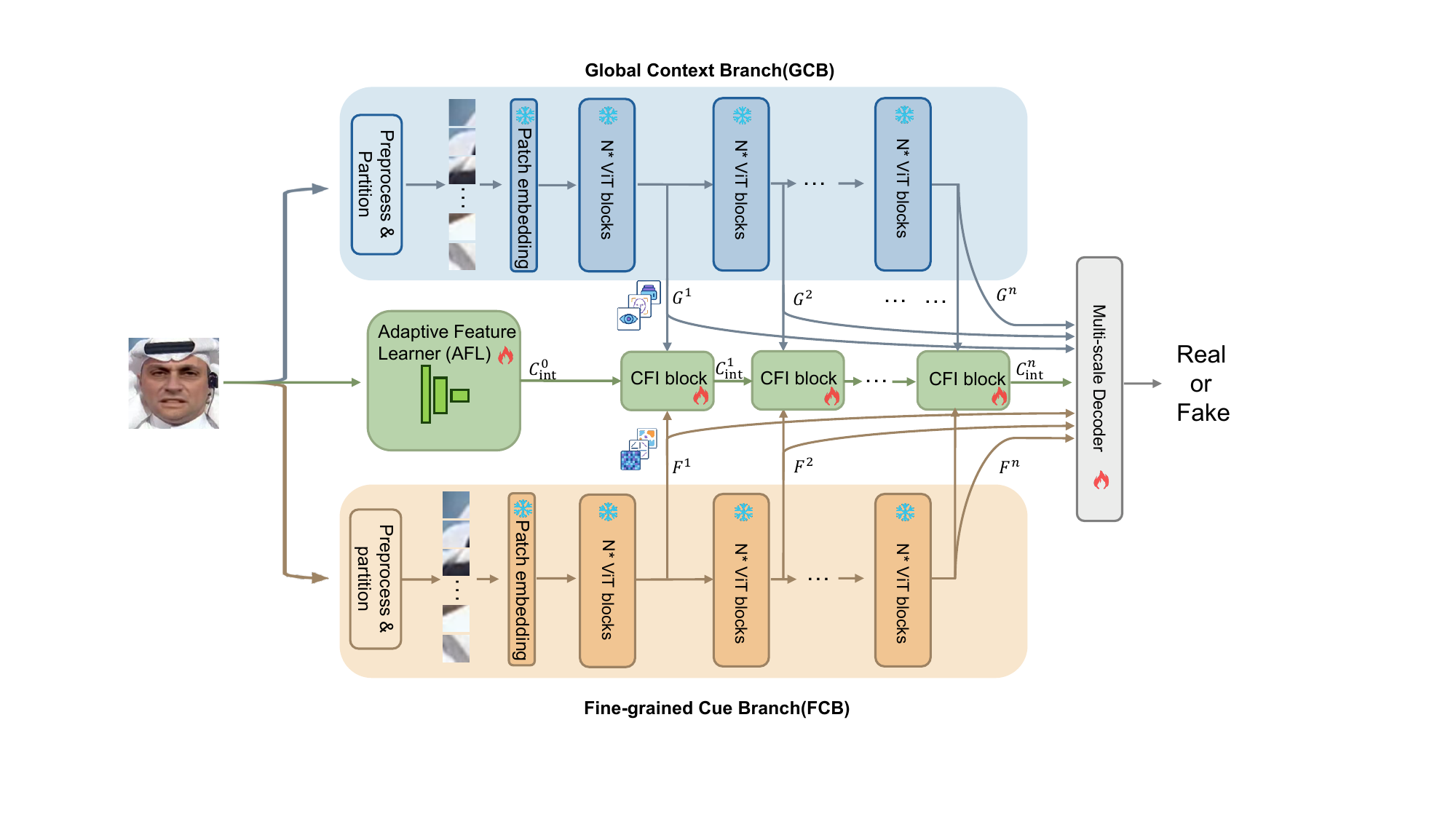}
\caption{Overview of the proposed framework. A Global Context Branch (GCB) and a Fine-grained Cue Branch (FCB) operate in parallel and are connected via adaptive cross-feature interaction modules, enabling effective collaboration between global context and fine-grained forgery cues.}

\label{overview}
\end{figure}

\begin{figure}[t]
\centering
\begin{minipage}[t]{0.32\linewidth}
\centering
\IfFileExists{AFL.pdf}
{\includegraphics[width=\linewidth]{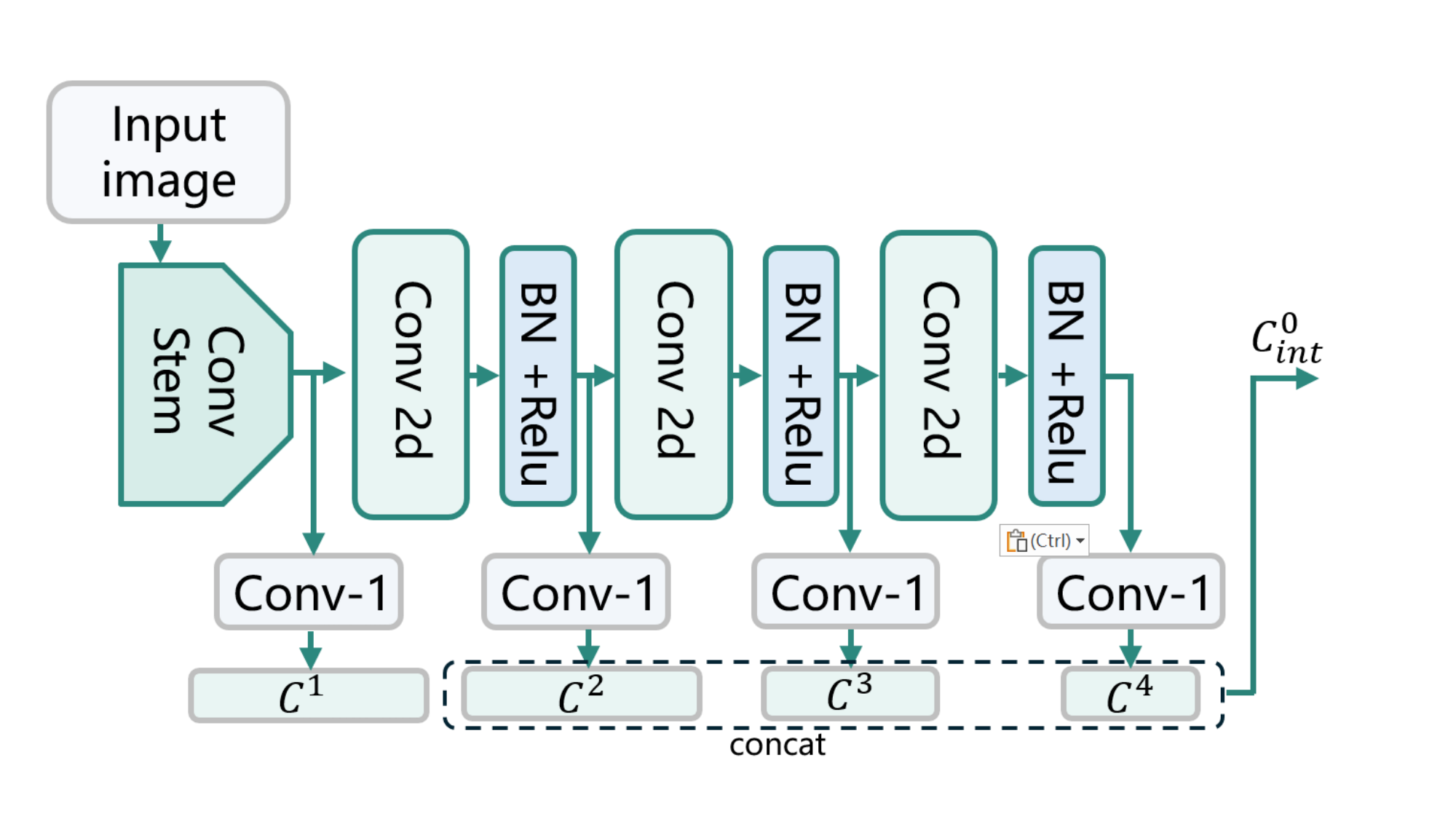}}
{\fbox{\parbox[c][0.18\textheight][c]{0.88\linewidth}{\centering AFL}}}
\vspace{2pt}
\small (a) Adaptive Feature Learner
\end{minipage}
\hfill
\begin{minipage}[t]{0.32\linewidth}
\centering
\IfFileExists{CFI.pdf}
{\includegraphics[width=\linewidth]{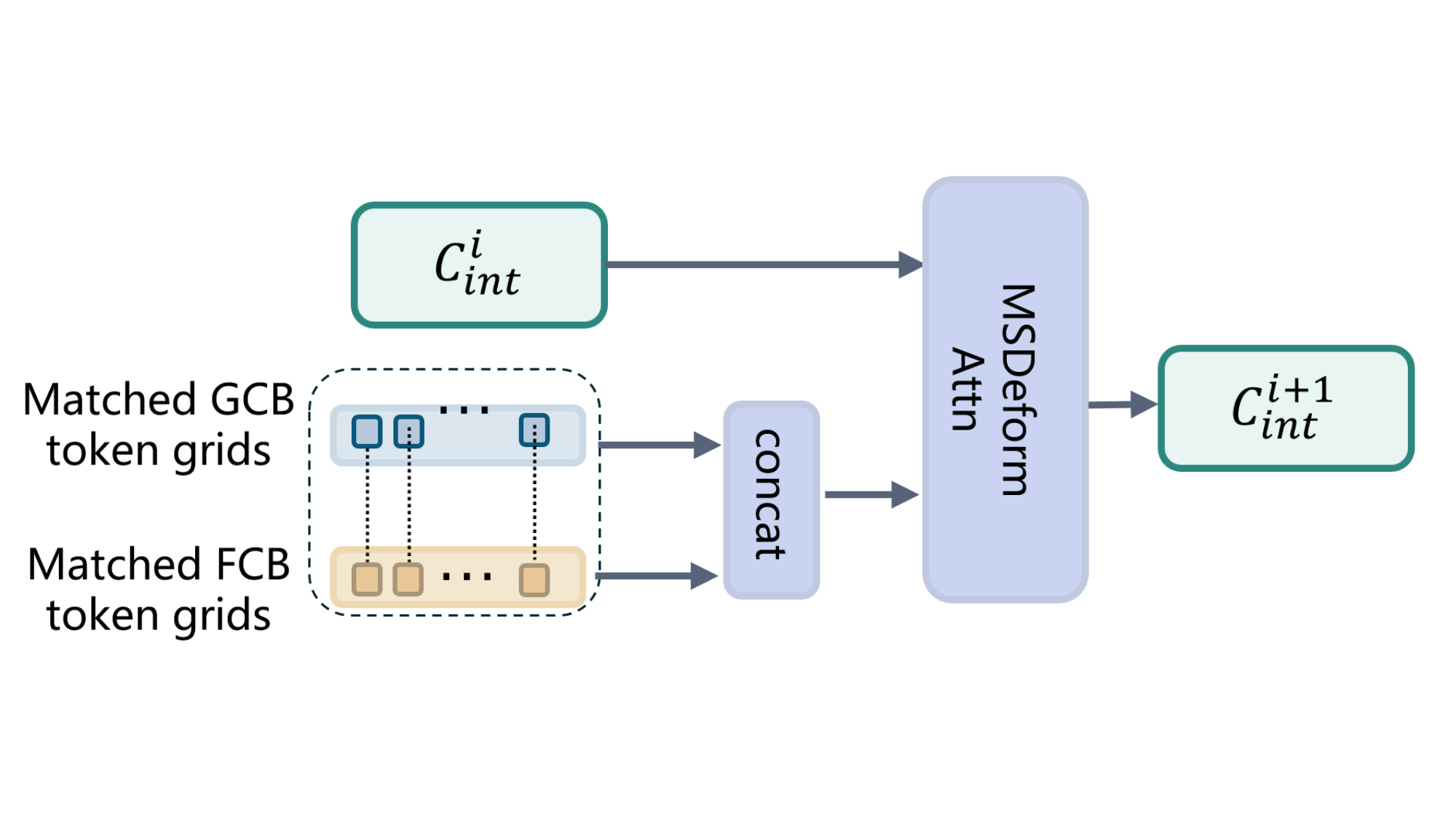}}
{\fbox{\parbox[c][0.18\textheight][c]{0.88\linewidth}{\centering CFI}}}
\vspace{2pt}
\small (b) Cross-Feature Interaction
\end{minipage}
\hfill
\begin{minipage}[t]{0.32\linewidth}
\centering
{\includegraphics[width=\linewidth]{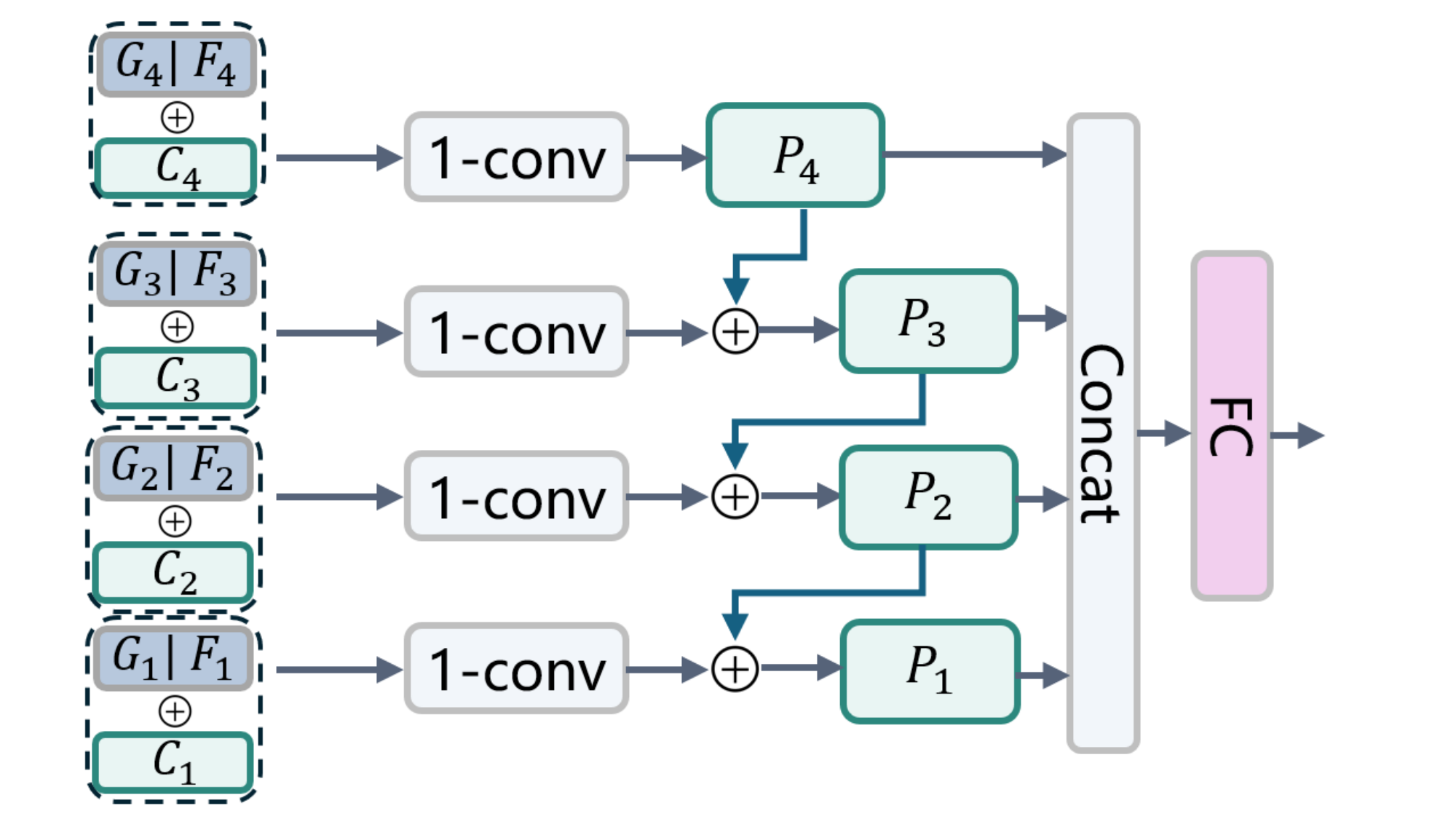}}
\vspace{2pt}
\small (c) Multi-scale Decoder
\end{minipage}
\caption{Detailed structures of the core components in the proposed framework. (a) The Adaptive Feature Learner (AFL) extracts multi-scale spatial priors. (b) The Cross-Feature Interaction (CFI) block progressively exchanges complementary information between global context and fine-grained forgery cues. (c) The multi-scale decoder aggregates hierarchical interaction features for the final real/fake prediction.}
\label{component_details}
\end{figure}

As illustrated in Fig.\ref{overview}, we propose a dual-branch framework for deepfake detection that explicitly models the complementarity between global context and fine-grained forgery cues. The framework consists of a Global Context Branch (GCB), a Fine-grained Cue Branch (FCB), and a set of adaptive cross-feature interaction modules that enable progressive and controlled information exchange across different representation levels. Fig.\ref{component_details} further presents the detailed structures of the three key components, including the Adaptive Feature Learner (AFL), the Cross-Feature Interaction (CFI) blocks, and the multi-scale decoder. To bridge the representational gap between the two branches, the AFL first extracts task-adaptive multi-scale spatial priors, and the CFI blocks then progressively refine these features by interacting with aligned global and fine-grained representations. Finally, the multi-scale decoder aggregates hierarchical features from both branches and produces the final real/fake prediction.
Different from conventional hybrid frameworks that mainly fuse task-specific local/global features or texture/noise cues, DBCF performs spatially aligned hierarchical fusion of complementary foundation-model representations. By aligning and progressively interacting CLIP-based global semantic features with DINOv3-based fine-grained structural features, the proposed framework enables effective collaboration between heterogeneous pretrained feature spaces.

\subsection{Global Context Branch}

A key challenge in deepfake detection is to achieve robust generalization across identities and manipulation methods while avoiding overfitting to localized, method-specific artifacts. To address this issue, the Global Context Branch (GCB) is introduced to extract stable, high-level semantic representations associated with identity, pose, and expression. These cues serve as global semantic priors for forgery detection. Since fine-grained artifact cues are often fragile under distribution shifts, whereas global semantics are more invariant, we adopt a frozen CLIP visual backbone as the core of the GCB. CLIP is pretrained with a large-scale multimodal contrastive objective, which naturally encodes strong semantic alignment and invariance. Freezing the backbone preserves these pretrained inductive biases and mitigates semantic collapse during training.

Specifically, following common adapter designs in pretrained ViT frameworks~\cite{simeoni2025dinov3,chen2022vision}, given an input image $x$, the ViT backbone of CLIP produces a sequence of intermediate token representations $\{H_g^{(l)}\}_{l=1}^{L}$, where $H_g^{(l)} \in \mathbb{R}^{N_{1} \times d_{1}}$ denotes the output of the $l$-th transformer layer, with $N_{1}$ being the number of tokens and $d_{1}$ being the token embedding dimension of the GCB. To construct multi-granular global context, we select a subset of layers indexed by $\mathcal{L}_g \subseteq \{1,\dots,L\}$ and define the global context features as
\begin{equation}
G = \{\, H_g^{(l)} \mid l \in \mathcal{L}_g \,\}.
\end{equation}

Following prior empirical studies on utilizing intermediate transformer representations for hierarchical feature construction~\cite{simeoni2025dinov3}, we select four representative layers from the CLIP visual encoder at different depths, i.e., $\mathcal{L}_g = \{5,12,18,24\}$. The resulting multi-granular global features are subsequently paired with the FCB representations in the Cross-Feature Interaction module.

\subsection{Fine-grained Cue Branch}

While global semantic representations provide robustness, many deepfake artifacts manifest as localized and fine-grained inconsistencies, such as subtle texture distortions or boundary artifacts. To explicitly capture such cues, the Fine-grained Cue Branch (FCB) is designed to extract artifact-sensitive local representations that complement the global context.

Specifically, fine-grained forgery cues refer to localized manipulation traces that are often weak in global semantics but evident in local visual patterns. Typical examples include blending boundaries, texture/color/illumination mismatches, local blurring or detail degradation, and subtle structural distortions in facial components. Since these cues are spatially localized and closely related to patch-level structure and appearance, we employ DINOv3 to construct the FCB. Benefiting from its self-supervised visual pretraining, DINOv3 preserves rich local structural representations and has demonstrated strong performance on dense prediction tasks such as semantic segmentation, indicating its capability to model spatially detailed visual information. This property makes it well suited for complementing the CLIP branch, which mainly captures global semantic and contextual information. Moreover, DINOv3 supports variable input resolutions, enabling flexible spatial alignment with the Global Context Branch (GCB).

To facilitate token-wise interaction between the two branches, we adopt a resolution-adaptive preprocessing strategy that ensures spatial alignment. Specifically, we resize the input image $x$ to a square resolution $S \times S$ with
\begin{equation}
S = \sqrt{N_1} \cdot P_{\mathrm{FCB}},
\end{equation}
where $P_{\mathrm{FCB}}$ denotes the patch size of the FCB backbone. With this choice, the token number produced by the FCB satisfies $N_2 = (S/P_{\mathrm{FCB}})^2 = N_1$. The resized image is then fed into the FCB, producing a sequence of intermediate token representations $\{H_f^{(l)}\}_{l=1}^{L}$, where $H_f^{(l)} \in \mathbb{R}^{N_2 \times d_2}$ and $d_2$ denotes the feature dimension of the FCB.

The fine-grained features extracted by the FCB are formulated as:
\begin{equation}
F = \{\, H_f^{(l)} \mid l \in \mathcal{L}_f \,\}, \quad \mathcal{L}_f = \{5,12,18,24\}.
\end{equation}

These fine-grained features provide localized and artifact-sensitive cues that complement the global context for forgery detection.

\subsection{Adaptive Feature Learner}
While the Global Context Branch (GCB) and Fine-grained Cue Branch (FCB) provide complementary representations, the frozen features limit adaptability to dataset-specific manipulation patterns. To enhance adaptability to downstream tasks, following common practice in recent forgery detection frameworks~\cite{shao2025deepfake,simeoni2025dinov3}, we incorporate a parameter-efficient Adaptive Feature Learner (AFL) as an auxiliary component to provide task-adaptive cues.

As shown in Fig.\ref{component_details}(a), the AFL is implemented as a trainable four-stage convolutional spatial prior module. It consists of a convolutional stem followed by three stride-2 convolutional stages, which progressively extract feature maps at $1/4$, $1/8$, $1/16$, and $1/32$ of the input resolution. Given an input image $x$, the AFL produces a hierarchy of multi-scale feature maps
$C^{(k)} \in \mathbb{R}^{H_k \times W_k \times C_k}$,
where the spatial resolution decreases and the semantic abstraction increases with $k$.
These multi-scale features provide complementary spatial cues and serve as auxiliary representations for subsequent cross-feature interaction.

To enable unified processing with transformer-based features, each feature map is first projected by a $1\times1$ convolution into a shared embedding space of dimension $d_c=d_1+d_2$. The three lower-resolution feature maps used for cross-feature interaction are augmented with a learnable level embedding $e^{(k)}$ and flattened into sequences of tokens:
\begin{equation}
\tilde{C}^{(k)}
=
\mathrm{Flatten}\!\left(\mathrm{Proj}_k(C^{(k)})\right)+e^{(k)}
\in \mathbb{R}^{N_k \times d_c},
\quad
N_k=H_kW_k,
\quad k=2,3,4.
\end{equation}

For cross-feature interaction with the GCB and FCB, the sequences from scales $k=2,3,4$ are concatenated along the token dimension to form
\begin{equation}
C_{\mathrm{int}}^{(0)}
=
[\tilde{C}^{(2)};\tilde{C}^{(3)};\tilde{C}^{(4)}]
\in \mathbb{R}^{N_{\mathrm{int}} \times d_c},
\qquad
N_{\mathrm{int}}=\sum_{k=2}^{4}N_k.
\end{equation}
The highest-resolution feature ${C}^{(1)}$  preserves its spatial structure and is reserved for the subsequent multi-scale decoding stage.

\subsection{Cross-Feature Interaction}

To integrate the multi-scale AFL features with spatially aligned global and local features, 
the Cross-Feature Interaction (CFI) module updates the interaction features via a cross-attention mechanism, as shown in Fig.\ref{component_details}(b). 
Let $G_s \in \mathbb{R}^{N_s \times d_1}$ and $F_s \in \mathbb{R}^{N_s \times d_2}$ denote the global and local token representations at the $s$-th selected representation level, respectively, where $s=1,\ldots,4$ corresponds to the selected transformer layers $\{5,12,18,24\}$. Since the input resolutions and patch sizes are chosen to yield spatially aligned token grids in the two branches, $G_s$ and $F_s$ are concatenated along the channel dimension.

Starting from the multi-scale AFL interaction features $C_{\mathrm{int}}^{(0)}$, each CFI module progressively updates them using the paired GCB--FCB features at the corresponding representation level:

\begin{equation}
\begin{aligned}
C_{\mathrm{int}}^{(s)}
={}& C_{\mathrm{int}}^{(s-1)}
+ \mathrm{MSDeformAttn}\Bigl(
    \mathrm{LayerNorm}(C_{\mathrm{int}}^{(s-1)}), \\
&\hspace{9.6em}
    \mathrm{LayerNorm}([G_s \mid F_s])
\Bigr),
\qquad s=1,\ldots,4.
\end{aligned}
\end{equation}

Here, $C_{\mathrm{int}}^{(0)}$ consists of the AFL features at $1/8$, $1/16$, and $1/32$ input resolutions. The deformable attention operation allows these AFL features with different token lengths to interact with the spatially aligned global and local representations. In this way, the CFI modules progressively incorporate complementary global and local cues into the task-adaptive multi-scale features.

\subsection{Multi-Scale Decoder and Training Objective}

As shown in Fig.\ref{component_details}(c), the multi-scale decoder aggregates the refined hierarchical features through a top-down fusion strategy. Given the AFL and CFI outputs, we split the final CFI output token sequence $C_{\mathrm{int}}$ into three groups and reshape them into spatial feature maps $C_2$, $C_3$, and $C_4$ at progressively lower resolutions. The highest-resolution feature map $C_1$ is directly taken from the shallow AFL feature. When CFI features are used, the interaction outputs from GCB and FCB are partitioned and resized to match the corresponding AFL feature maps and added to them. After normalization, the preceding fusion process produces four spatial feature maps $\{C_i\}_{i=1}^{4}$ at $1/4$, $1/8$, $1/16$, and $1/32$ input resolutions, respectively. The decoder progressively fuses these features to generate the final prediction.

Each feature map is first projected to a unified channel dimension $d_o$ through a $1\times1$ convolution:
\[
P_i = \mathrm{Conv}_{1\times1}(C_i), \quad i = 1, \dots, 4,
\]
where $P_1$ and $P_4$ denote the highest and lowest resolution decoder features, respectively.

The decoder follows a top-down fusion strategy, where higher-level features are progressively upsampled and merged with lower-level ones. Specifically, the lowest-resolution projected feature is retained as $\tilde{P}_4=P_4$, and the remaining levels are progressively fused as:
\[
\tilde{P}_4 = P_4, \qquad
\tilde{P}_i =
P_i + \mathrm{Up}(\tilde{P}_{i+1}),
\quad i=3,2,1.
\]
Each fused representation is then transformed into a scale-specific global descriptor:
\[
g_i = \mathrm{GAP}(\tilde{P}_i), \quad i=1,2,3,4,
\]
with $\mathrm{Up}(\cdot)$ denoting bilinear interpolation and $\mathrm{GAP}(\cdot)$ global average pooling.

The final image-level representation is obtained by concatenating all scale-specific descriptors and feeding it into a linear classifier to predict the two-class logits:
\[
\mathbf{z} = \mathrm{FC}\Big( [g_1; g_2; g_3; g_4] \Big)
\in \mathbb{R}^{2},
\]
Finally, the model is optimized using the cross-entropy loss:
\[
\mathcal{L}_{\mathrm{cls}}
=
-
\log
\frac{\exp(z_y)}
{\sum_{c=0}^{1}\exp(z_c)},
\]
where $y \in \{0,1\}$ denotes the ground-truth label.

\section{Experiments}
\subsection{Settings}
\subsubsection{Datasets}

To comprehensively assess the effectiveness and generalization capability of the proposed method, we conduct experiments under both cross-dataset and cross-manipulation evaluation protocols. For cross-dataset evaluation, we follow the widely adopted setting in which the model is trained on one dataset and tested on several unseen datasets. Specifically, we use the c23 compressed version of FaceForensics++ (FF++)~\cite{rossler2019faceforensics++} for training, which consists of 1,000 pristine videos and 4,000 manipulated videos generated by four manipulation techniques. Following the official split of FF++, only its training subset, consisting of 720 pristine videos and 2,880 manipulated videos, is used for training. No cross-validation or test-time adaptation is involved in the cross-dataset evaluation. The trained model is then evaluated on three challenging benchmarks: Celeb-DF v2~\cite{li2020celeb}, DFDC~\cite{dolhansky2020deepfake}, DFDCP~\cite{dolhansky2019deepfake}, covering diverse real-world conditions and distribution shifts. Following the commonly adopted evaluation protocol for fair cross-dataset comparison, we evaluate our method on the official test split of each target benchmark. To further investigate robustness against unseen forgery types, we adopt DF40~\cite{yan2024df40}, a comprehensive dataset containing 40 manipulation techniques spanning a broad range of facial forgery categories, including face swapping, facial reenactment, and full-face synthesis.

\subsubsection{Evaluation Metrics}
We report both frame-level and video-level Area Under the ROC Curve (AUC) to accommodate different evaluation granularities. Following standard practice, video-level AUC is computed by averaging the predicted probabilities over all sampled frames within each video. For extensive face anti-spoofing (FAS) experiment, we additionally employ the Half Total Error Rate (HTER) to further measure cross-domain generalization performance.

\subsubsection{Implementation Details}
We adopt CLIP-ViT-L/14-336 as the backbone for global feature extraction and DINOv3-ViT-L/16 to provide complementary fine-grained representations. AFL uses 64 base convolutional channels and projects the multi-scale features to a shared dimension of 2048. The decoder projection dimension is set to 512. During training, we sample 8 frames from each video. For the cross-dataset comparisons in Tables~\ref{tab:deepfake_framelevel} and~\ref{tab:deepfake_videolevel}, we sample 32 frames from each test video to match the evaluation setting adopted by the reported baseline results. All frames are aligned using RetinaFace and resized to 224$\times$224. Consistent with the spatial-alignment procedure described in Section~3, the preprocessed frames are subsequently resized to 336$\times$336 for the CLIP-based GCB and 384$\times$384 for the DINOv3-based FCB before being fed into the respective backbones. We use only image (patch) tokens for cross-branch alignment, excluding the CLS token. With patch sizes of 14 and 16, respectively, both branches produce a $24\times24$ grid containing 576 image tokens. Following other existing works, several image augmentations are introduced during training, including random Brightness Contrast, Image Compression, and the SBI-based augmentation strategy~\cite{shiohara2022detecting}. No additional data augmentation is applied during testing. All experiments are implemented within PyTorch framework and conducted on a single NVIDIA A100 GPU. All models are trained for 10 epochs. We use the Adam optimizer with a fixed learning rate of $2 \times 10^{-4}$ for all trainable parameters. The batch size is set to 32 across all experiments to ensure fair comparison. For repeated-run experiments, we perform three independent runs using different fixed random seeds and report the mean AUC and the corresponding standard deviation across these runs.
\subsection{Main Results}

\subsubsection{Cross-dataset evaluation}

\begin{table}[!t]
\centering
\renewcommand{\arraystretch}{0.9}

\caption{Cross-dataset comparison using frame-level AUC. The results reported in the table are taken from~\cite{yan2023deepfakebench,ma2026specificity}, or directly obtained from the corresponding original papers.}
\label{tab:deepfake_framelevel}
\begin{tabular}{c|c|ccc}
\hline
\textbf{Method} & \textbf{Venue} & \multicolumn{1}{c}{\textbf{CDF}} & \multicolumn{1}{c}{\textbf{DFDC}} & \textbf{DFDCP} \\ \hline
{Xception~\cite{rossler2019faceforensics++}}         & ICCV'19 & 73.65 & 70.77 & 73.74 \\
{FaceX-ray~\cite{li2020face}}        & CVPR'20 & 67.86 & 63.26 & 69.42 \\
{RECCE~\cite{cao2022end}}            & CVPR'22 & 73.19 & 71.33 & 74.19 \\
{SBI~\cite{shiohara2022detecting}}              & CVPR'22 & 81.30 & 71.96 & 79.90 \\
{UIA-ViT~\cite{zhuang2022uia}}          & ECCV'22 & 82.41 & --   & 75.80  \\
{UCF~\cite{yan2023ucf}}              & ICCV'23 & 75.27 & 71.91 & 75.94 \\
{ProDet~\cite{cheng2024can}}           & NIPS'24 & 84.48 & 72.40 & 81.16 \\
{LSDA\cite{yan2024transcending}}             & CVPR'24 & 83.00 & 73.60 & 81.50 \\
{DeepFake-Adapter~\cite{shao2025deepfake}} & IJCV'25 & 71.74 & 72.66 & --    \\
{FIA-USA~\cite{ma2026specificity}}          & NIPS'25 & \textbf{86.70} &  --   & 81.80  \\ 
{CRDA}~\cite{chou2026improving} & AAAI'26        & \underline{85.36}         & 74.29         & 79.73           \\ \hline
DBCF (Ours)     &  --     & 83.71$\pm$0.62 & \underline{78.02$\pm$0.22} & \underline{85.90$\pm$0.86} \\ 
DBCF (Ours) + SBI    &  --     & 83.25$\pm$0.57 & \textbf{80.08$\pm$0.30} & \textbf{88.66$\pm$0.16} \\ \hline
\end{tabular}
\end{table}

\begin{table}[!t]
\centering 
\renewcommand{\arraystretch}{0.9}
\caption{Comparison with SOTA methods using the video-level AUC. Results marked with * are obtained by using the authors' released models or code, while the remaining results are directly taken from ~\cite{yan2023deepfakebench} or corresponding original papers.}
\label{tab:deepfake_videolevel}
\begin{tabular}{c|c|ccc}
\hline
\textbf{Method}         & \textbf{Venue} & \textbf{CDF} & \textbf{DFDC} & \textbf{DFDCP} \\ \hline
{FaceX-ray~\cite{li2020face}}      & CVPR'20        &  --          & --            & 71.1           \\
{FTCN~\cite{zheng2021exploring}}           & ICCV'21        & 86.9         & 67.6          & 74.0           \\
{SBI~\cite{shiohara2022detecting}}            & CVPR'22        & \underline{92.8}         & 71.9          & 85.5           \\
{UIA-ViT*~\cite{zhuang2022uia}}       & ECCV'22        & 82.4         & 75.0          & 75.8           \\
{CFM*~\cite{luo2023beyond}}           & TIFS'23        & 85.3         & 75.0          & 80.2           \\
{AltFreezing*~\cite{wang2023altfreezing}}   & CVPR'23        & 85.1         & 71.7          & 79.3           \\
{LSDA~\cite{yan2024transcending}}           & CVPR'24        & 91.1         & 77.0          & 81.2           \\
{NACO~\cite{alshehri2025deep}}           & AJSE'25        & 89.5         & 76.7          & --             \\
{SDR~\cite{chu2025reduced}}        & ICASSP'25        & 88.5         & 76.2          & --          \\ 
{M2F2-Det*~\cite{guo2025rethinking}}        & CVPR'25        & 83.2         & 76.4          & 71.2          \\ 
{FIA-USA~\cite{ma2026specificity}}        & NIPS'25        & \textbf{94.1}         & 73.2          & 86.6           \\  \hline
DBCF (Ours)  &          --      & 88.6$\pm$0.4         & \underline{80.9$\pm$0.3}          & \underline{87.7$\pm$0.6}           \\ 
DBCF (Ours) + SBI  &          --      & 89.2$\pm$0.2         & \textbf{82.9$\pm$0.2}          & \textbf{92.5$\pm$0.1}          \\ \hline
\end{tabular}
\end{table}

We first evaluate the generalization ability of our method by comparing frame-level Area Under the ROC Curve (AUC) across several unseen datasets, including Celeb-DF v2, DFDC, and DFDCP, which are completely excluded during training.
Table \ref{tab:deepfake_framelevel} presents the cross-dataset results of our method alongside several representative baselines. Overall, our approach demonstrates strong generalization across all benchmarks. While some methods achieve slightly higher performance on CDF, DBCF attains the highest performance on DFDC and DFDCP, demonstrating strong overall generalization across the evaluated target datasets. The variation across target datasets can be attributed to their different degrees of distribution shift and manipulation diversity. Although CDF contains high-quality and visually realistic forged videos, its manipulation distribution is relatively concentrated, and thus transferable forgery patterns may remain comparatively consistent. In contrast, DFDC contains more diverse identities, backgrounds, acquisition conditions, and manipulation algorithms. Such real-world variations may obscure subtle local forgery traces, making DFDC more challenging, as reflected by the lower frame-level AUC of 78.02 compared with other datasets. We further investigate its compatibility with data-centric enhancement strategies. The results demonstrate that DBCF can be seamlessly combined with existing data augmentation methods, such as SBI~\cite{shiohara2022detecting}, leading to performance improvements across multiple datasets. These findings suggest that DBCF offers a solid foundation for cross-dataset deepfake detection and shows potential for further improvement when combined with complementary data augmentation techniques during training.

We also report video-level AUC in Table \ref{tab:deepfake_videolevel}. The video-level results exhibit trends consistent with the frame-level evaluation, while showing improved stability across datasets. This indicates that the proposed method produces temporally consistent predictions across frames, and that video-level aggregation helps mitigate the impact of noisy frame-level predictions, particularly on challenging datasets such as DFDC.

\subsubsection{Cross-manipulation evaluation}

\begin{table}[t]
\centering

\caption{Cross-manipulation comparison on five representative face swapping forgery types in DF40~\cite{yan2024df40} using frame-level AUC (\%)}
\resizebox{\textwidth}{!}{
\begin{tabular}{@{}c| c| c c c c c c@{}}
\hline
\textbf{Method} & \textbf{Venue} & \textbf{uniface} & \textbf{facedancer} & \textbf{fsgan} & \textbf{inswap} & \textbf{simswap} & \textbf{Avg.} \\
\hline
RECCE~\cite{cao2022end}   & CVPR'22  & 84.2 & 78.3 & 88.4 & 79.5 & 73.0 & 80.7 \\
SBI~\cite{shiohara2022detecting}     & CVPR'22  & 64.4 & 44.7 & 87.9 & 63.3 & 56.8 & 63.4 \\
IID~\cite{huang2023implicit}     & CVPR'23  & 79.5 & 79.0 & 86.4 & 74.4 & 64.0 & 76.7 \\
UCF~\cite{yan2023ucf}     & ICCV'23  & 78.7 & 80.0 & 88.1 & 76.8 & 64.9 & 77.7 \\
LSDA~\cite{yan2024transcending}    & CVPR'24  & 85.4 & 75.9 & 83.2 & 81.0 & 72.7 & 79.6 \\
CDFA~\cite{lin2024fake}    & ECCV'24  & 76.5 & 75.4 & 84.8 & 72.0 & 76.1 & 77.0 \\
ProDet~\cite{cheng2024can} & NeurIPS'24 & 84.5 & 73.6 & 86.5 & 78.8 & 77.8 & 80.2 \\
FIA-USA~\cite{ma2026specificity} & NIPS'25  & 91.8 & 83.0 & 86.3 & 87.4 & 91.0 & 87.8 \\
\hline
DBCF (Ours)    & -- & \underline{97.7$\pm$0.3} & \underline{90.0$\pm$0.3} & \underline{95.4$\pm$0.2} & \underline{92.3$\pm$0.8} & \underline{93.4$\pm$1.3} & \underline{93.8$\pm$0.4} \\
DBCF (Ours) + SBI   & --   & \textbf{97.9$\pm$0.5} & \textbf{91.1$\pm$0.5} & \textbf{95.3$\pm$0.4} & \textbf{93.1$\pm$1.9} & \textbf{94.8$\pm$0.4} & \textbf{94.4$\pm$0.4} \\
\hline
\end{tabular}}
\label{tab:df40_framelevel}
\end{table}

To further examine the robustness of our method against unseen manipulation types, we conduct a cross-manipulation evaluation on the DF40 benchmark~\cite{yan2024df40}, which contains five distinct generation pipelines: uniface, facedancer, fsgan, inswap, and simswap. As reported in Table \ref{tab:df40_framelevel}, our method obtains higher AUC than the listed baselines on the five evaluated manipulation pipelines and achieves the highest average AUC in this comparison. Nevertheless, noticeable performance differences can be observed across manipulation methods. In particular, facedancer and simswap are relatively more challenging than uniface and fsgan. This may be because modern high-fidelity face-swapping pipelines are designed to better preserve facial attributes and reduce visible blending inconsistencies, leaving weaker and more localized forgery traces for detection. In contrast, manipulation methods that leave more visible and straightforward forgery artifacts are comparatively easier to identify.

A comparative examination of Tables \ref{tab:deepfake_framelevel}--\ref{tab:df40_framelevel} reveals that many existing deepfake detection methods may achieve strong performance on certain datasets, but their effectiveness often fails to generalize to more diverse and complex forgery patterns in the DF40 cross-manipulation setting.
This indicates that prior approaches might exploit dataset-specific or manipulation-dependent cues rather than capturing intrinsic forgery characteristics. In contrast, our method consistently achieves strong performance across all unseen types. Moreover, its improvements are particularly evident on relatively challenging manipulation methods, indicating improved robustness to high-fidelity forgeries with less conspicuous artifacts. We attribute this robustness to the use of more comprehensive and complementary representations, which allow the model to focus on fundamental forgery traits that remain stable despite significant shifts in manipulation strategies and generation mechanisms. 

\subsubsection{Ablation Study}

Table~\ref{tab:ablation1} presents a comprehensive ablation study evaluating the individual and joint contributions of the Global Context Branch (GCB), the Fine-grained Cue Branch (FCB), and the Multi-granular feature aggregation module (MG). Using either GCB or FCB alone yields reasonable performance, indicating that both global semantic consistency and fine-grained artifacts are informative for forgery detection. However, directly concatenating the outputs of GCB and FCB (without MG) does not consistently improve performance and even degrades results on some datasets. This suggests that simple concatenation of heterogeneous features is insufficient for effective integration. The performance drop may stem from directly merging features with different distributions, which can lead to collapsed or suboptimal feature patterns.
Introducing MG consistently improves performance by leveraging hierarchical representations instead of relying solely on final-layer tokens. When combined with both branches, our Adaptive Feature Learner and Cross-Feature Interaction mechanism further enhance performance, demonstrating that spatially aligned and interaction-aware fusion is critical for effectively integrating global and local cues.
Enabling all components achieves the best results across datasets, validating the effectiveness of our multi-granular and structured fusion design for generalizable deepfake detection.

\begin{table}[]
\centering
\caption{Ablation study of different components on frame-level AUC across multiple datasets. GCB denotes the Global Context Branch for modeling global semantic consistency, FCB denotes the Fine-grained Cue Branch for capturing local manipulation artifacts, and MG denotes the multi-granular feature aggregation module for hierarchical feature fusion.}
\begin{tabular}{ccc|ccccc}
\hline
\multicolumn{3}{c|}{\textbf{Component Settings}} & \multicolumn{5}{c}{\textbf{Frame-level AUC}}                                    \\ \hline
GCB       & MG     & FCB     & FF++ & Celeb-DF & DFDC & facedancer & inswap \\
\ding{51} &        &         & 82.6$\pm$0.1     & 75.6$\pm$1.0     & 72.6$\pm$0.6 & 77.4$\pm$0.6            & 76.2$\pm$0.9        \\
          &        & \ding{51}    & 92.5$\pm$0.3     & 82.8$\pm$0.3     & 73.2$\pm$0.2 & 75.0$\pm$0.3           & 76.1$\pm$0.6        \\
\ding{51} &        &  \ding{51}   & 88.8$\pm$0.2     & 75.7$\pm$0.7     & 73.2$\pm$0.5 & 78.0$\pm$0.5            & 79.4$\pm$1.1        \\
          &\ding{51} & \ding{51}  & 96.0$\pm$0.1     & 80.7$\pm$0.4      & 76.8$\pm$0.9   & 88.1$\pm$0.4            & 92.2$\pm$1.4         \\
\ding{51} &\ding{51} &            & 95.9$\pm$0.2      & 81.7$\pm$0.9     & 75.3$\pm$0.5  & 83.9$\pm$0.6             & 92.3$\pm$1.2         \\

\ding{51} &\ding{51} & \ding{51}  & \textbf{96.3$\pm$0.1}     & \textbf{83.7$\pm$0.6}    & \textbf{78.0$\pm$0.2} & \textbf{90.0$\pm$0.3}            & \textbf{92.3$\pm$0.8}      \\ \hline
\end{tabular}
\label{tab:ablation1}
\end{table}

To further assess the effectiveness-efficiency trade-off of the proposed dual-branch framework, we compare four representative variants in Table~\ref{tab:efficiency}. The average frame-level AUC is computed over the five evaluation settings reported in Table~\ref{tab:ablation1}. Consistent with the component-wise observations above, DBCF achieves the best average performance among all evaluated variants, confirming that the gain provided by the dual-branch framework is substantial rather than marginal. In terms of efficiency, DBCF requires 56 ms per frame, corresponding to approximately 17.9 FPS, with 678M parameters and 544 GFLOPs. Although incorporating complementary foundation models introduces additional computational overhead, the resulting inference speed remains acceptable under frame-sampled deepfake detection. Therefore, the improved generalization capability represents a reasonable trade-off against the increased latency and computational cost.
More specifically, compared with naive concatenation, DBCF improves the average frame-level AUC from 79.03\% to 88.06\%, while increasing inference latency from 33 ms to 56 ms per frame and computational cost from 367 to 544 GFLOPs. These results explicitly quantify the additional computational cost associated with the performance improvement of DBCF.

\begin{table}[t]
\centering
\small
\caption{Effectiveness-efficiency comparison of representative model variants. Latency and computational metrics are measured under the same inference setting and reported on a per-frame basis.}
\begin{tabular*}{\linewidth}{@{\extracolsep{\fill}}lcccc@{}}
\hline
\textbf{Variant} & \makecell{\textbf{Avg. frame-level} \\ \textbf{AUC} $\uparrow$} & \makecell{\textbf{Latency} \\ \textbf{(per frame)}} & \makecell{\textbf{Params}} & \makecell{\textbf{GFLOPs}} \\
\hline
GCB only                  & 76.83          & 30 ms & 330M & 248 \\
FCB only                 & 79.92          & 31 ms & 330M & 249 \\
GCB + FCB (Naive concat)           & 79.03          & 33 ms & 607M & 367 \\
DBCF (Ours)                            & \textbf{88.06} & 56 ms & 678M & 544 \\
\hline
\end{tabular*}
\label{tab:efficiency}
\end{table}

\subsubsection{Visualization}
\begin{figure}[t]
\centering
\includegraphics[width=1.0\linewidth, trim=50 120 120 50, clip]{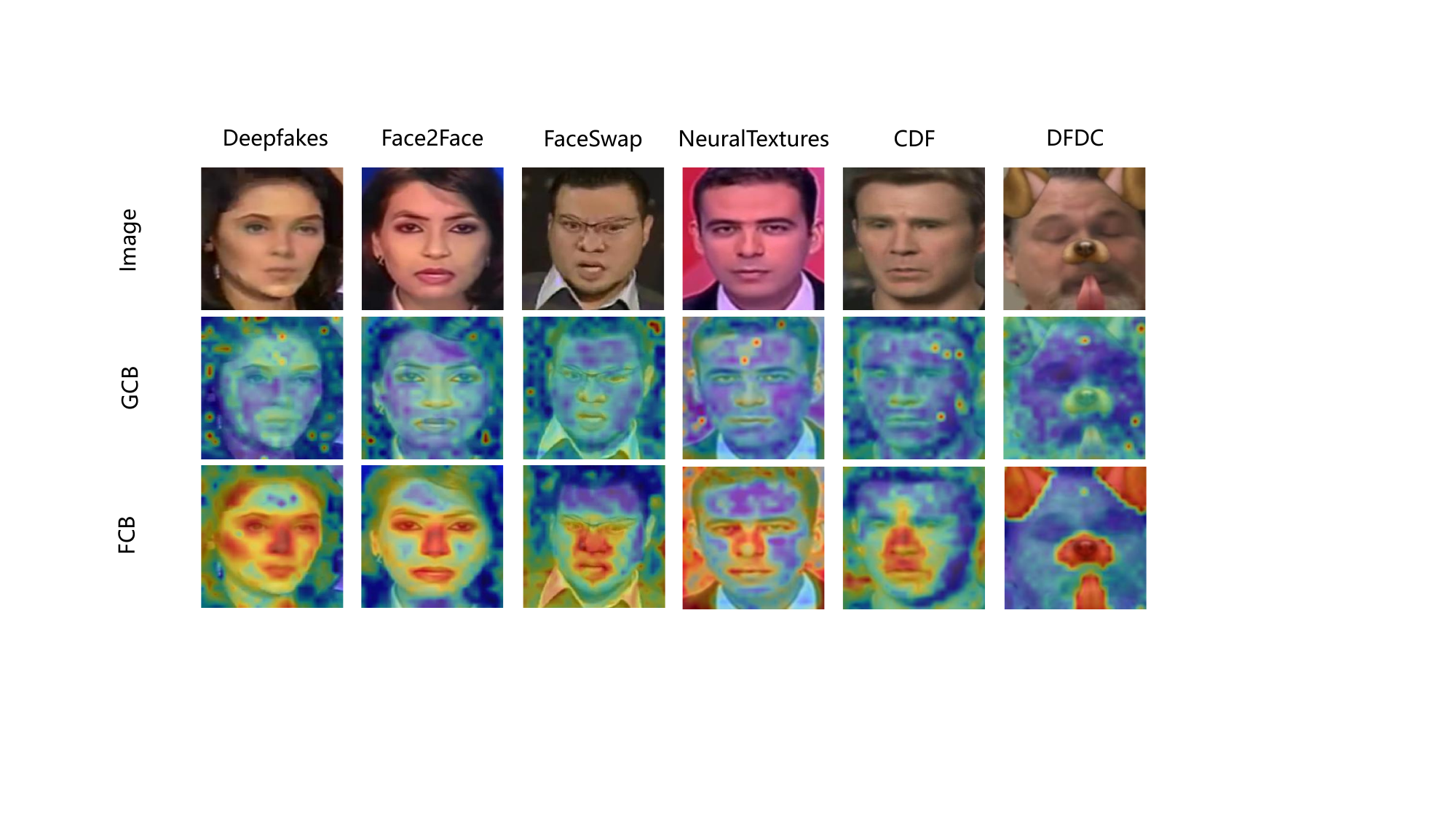}
\caption{Qualitative visualization of attention maps from the GCB and FCB branches on different forgery datasets.}
\label{Visualize}
\end{figure}

To qualitatively examine the complementary behavior of GCB and FCB, we visualize the branch-wise averaged attention maps in Fig.\ref{Visualize}. The GCB maps show relatively dispersed responses over the face and nearby regions, while the FCB maps exhibit more spatially coherent and locally concentrated responses within facial regions. This difference suggests that GCB tends to capture broader semantic/contextual cues, whereas FCB is more sensitive to localized appearance patterns, which is consistent with their complementary roles in deepfake detection.

\subsubsection{Robustness Analysis}
\begin{figure}[t]
\centering
\includegraphics[width=1.0\linewidth, trim=0 0 0 0, clip]{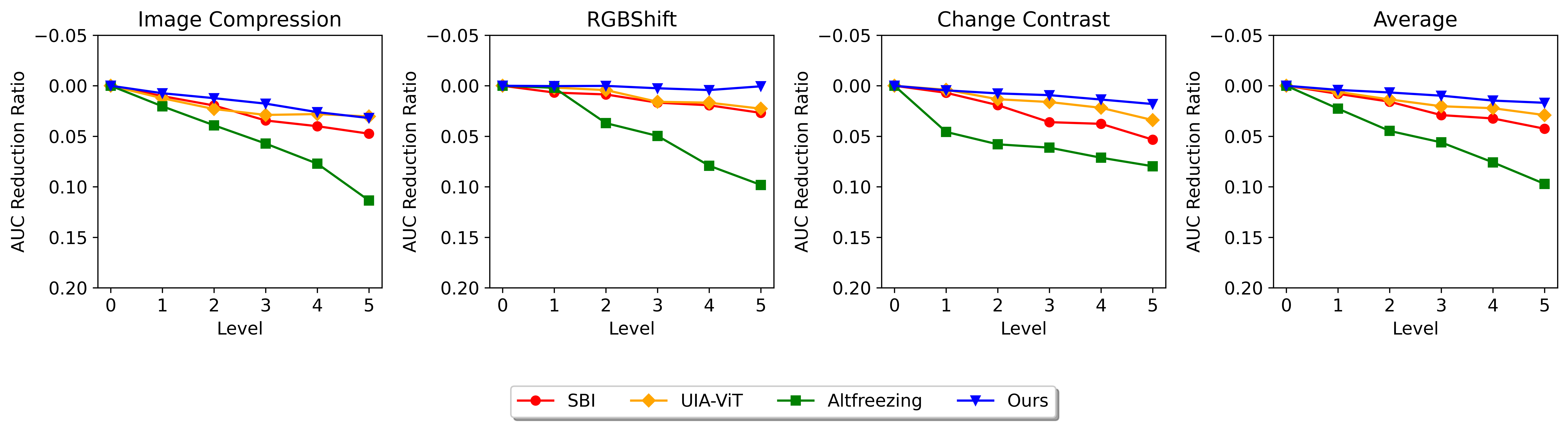}
\caption{Frame-level AUC reduction ratio under different degradation levels and perturbation types. "Average" score represents the mean across all levels for each type of perturbation.}
\label{robustness}
\end{figure}

Following AltFreezing~\cite{wang2023altfreezing}, we evaluate the robustness of different methods under common image perturbations in real-world scenarios, including image compression, RGB shift, and contrast variation. We report the frame-level AUC reduction ratio, defined as 
$(\text{AUC}_{\text{raw}} - \text{AUC}) / \text{AUC}_{\text{raw}}$, 
where $\text{AUC}_{\text{raw}}$ denotes the performance on clean (non-degraded) samples. A smaller value indicates stronger robustness. As shown in Fig.~\ref{robustness}, our method consistently exhibits the lowest AUC reduction across most perturbations, highlighting its robustness. While image compression causes a moderate drop due to the loss of subtle texture cues, performance under RGB shift and contrast variation remains largely unaffected.

\subsection{Extensive Experiments on Face Anti-Spoofing (FAS)}

\begin{table}[htb!]
\centering
\footnotesize
\setlength{\tabcolsep}{5pt}
\caption{Performance comparison of different face anti-spoofing methods under cross-dataset evaluation. All results for the compared methods are taken from~\cite{long2024generalized}. The metrics reported include HTER and AUC for each source-target dataset combination, 
as well as the mean performance across all settings. }
\begin{tabular}{c|cccc|cc}
\hline
\textbf{Method} &
\makecell{O\&C\&I$\rightarrow$M \\ HTER$\downarrow$/AUC$\uparrow$} &
\makecell{O\&M\&I$\rightarrow$C \\ HTER$\downarrow$/AUC$\uparrow$} &
\makecell{O\&C\&M$\rightarrow$I \\ HTER$\downarrow$/AUC$\uparrow$} &
\makecell{C\&I\&M$\rightarrow$O \\ HTER$\downarrow$/AUC$\uparrow$} &
\makecell{Avg \\ HTER$\downarrow$} &
\makecell{Avg \\ AUC$\uparrow$} \\
\hline
NAS-FAS & 19.53 / 88.63 & 16.54 / 90.18 & 14.51 / 93.84 & 13.80 / 93.43 & 16.10 & 91.52 \\
SSAN-R  & 6.67 / \textbf{98.75} & 10.00 / 96.67 & 8.88 / 96.79 & 13.72 / 93.63 & 9.82 & 96.46 \\
PatchNet& 7.10 / 98.46 & 11.33 / 94.58 & 13.40 / 95.67 & 11.82 / 95.07 & 10.91 & 95.95 \\
SA-FAS  & 5.95 / 96.55 & 8.78 / 95.37 & 6.58 / 97.54 & 10.00 / 96.23 & 7.83 & 96.42 \\
AG-FAS  & \underline{5.71} / 98.03 & 5.44 / 98.55 & 6.71 / 98.23 & 9.43 / 96.62 & 6.82 & 97.86 \\ \hline
Ours (GCB) & 5.95 / 98.50 & 2.67 / 99.54 & 17.14 / 91.24 & 10.14 / 96.31 & 8.98 & 96.40 \\
Ours (FCB) & \textbf{4.52} / 98.54 & \textbf{1.22} / \textbf{99.96} & \textbf{3.71} / \textbf{99.59} & \underline{5.99} / \underline{98.68} & \underline{3.86} & \underline{99.19} \\
Ours (Dual)& \underline{5.71} / \underline{98.64} & \underline{1.33} / \underline{99.85} & \underline{4.86} / \underline{99.29} & \textbf{2.92} / \textbf{99.66} & \textbf{3.71} & \textbf{99.36} \\
\hline
\end{tabular}

\label{tab:FAS}
\end{table}
We further evaluate the effectiveness of our framework on the face anti-spoofing task under cross-dataset settings. Experiments are conducted on four widely used benchmarks, and each source-target combination follows the standard Leave-One-Out protocol. As shown in Table \ref{tab:FAS}, we compare our method with several state-of-the-art approaches, including NAS-FAS~\cite{yu2020fas}, SSAN-R~\cite{wang2022domain}, PatchNet~\cite{wang2022patchnet}, SA-FAS~\cite{sun2023rethinking}, and the more recent AG-FAS~\cite{long2024generalized}.
Our proposed DBCF achieves consistently strong performance across all transfer scenarios in terms of both HTER and AUC. Specifically, the dual-branch model (Ours dual) maintains HTERs below 5 for all domain-shift settings, with AUC up to 99.66, outperforming the compared methods in average metrics (Mean HTER = 3.71, Mean AUC = 99.36).

We also conduct ablation studies to investigate the contributions of the global and local feature branches. Using only the GCB branch achieves low HTER in certain settings (e.g., 2.67\% on O\&M\&I $\rightarrow$ C) but suffers in others (17.14 on O\&C\&M $\rightarrow$ I), while the FCB branch alone improves overall HTER stability. The dual-branch design effectively combines the strengths of both branches, achieving robust performance across all cross-dataset scenarios.
Together, these results support the utility of combining global and local representations for the evaluated face forgery and spoofing detection tasks.

\section{Conclusion}
In this work, we proposed a dual-branch framework, namely DBCF, for detecting facial forgeries and spoofing attempts. It captures both global and local manipulation cues from manipulated images, enabling the model to learn more comprehensive and discriminative representations for forgery detection. Motivated by the complementary strengths of pretrained visual models, DBCF leverages CLIP’s strong semantic-level generalization while incorporating DINO’s fine-grained local features, allowing the framework to benefit from both high-level transferable semantic knowledge and subtle local artifact perception. This combination is particularly important for facial forgery and spoofing detection, where both overall semantic consistency and fine-grained manipulation traces contribute to reliable prediction. To effectively fuse these complementary cues without disrupting the pretrained feature patterns, we introduce a multi-scale spatial alignment mechanism that enables comprehensive learning of manipulation representations. By promoting more effective interaction and alignment across heterogeneous features at different spatial levels, the proposed mechanism improves the integration of global and local information while preserving the intrinsic strengths of each branch. Extensive experiments under cross-dataset and cross-manipulation settings show that our method achieves the best performance on most of the evaluated unseen datasets and manipulation types, demonstrating its generalization capability and robustness under distribution shifts. These results verify the effectiveness of the proposed design and suggest that combining heterogeneous pretrained representations is a promising direction for open-domain forgery detection. Overall, our framework provides a flexible foundation for open-domain detection tasks and offers potential for future extensions to video-level and multi-modal detection scenarios.

\section{Limitations}

Despite its strong generalization performance, DBCF has several practical limitations. Its complementary representation relies on two large pretrained backbones, CLIP-ViT-L and DINOv3-ViT-L. Although both backbones are frozen during training, jointly running them increases model size, computational cost, inference latency, and GPU memory demand compared with single-backbone alternatives. Specifically, DBCF requires 56 ms for single-frame inference, corresponding to approximately 17.9 FPS, with a peak GPU memory consumption of approximately 6.1 GB under the same inference setting. While these costs are accompanied by the performance gains reported in our evaluations, they may still restrict deployment in strict real-time or resource-constrained scenarios. Moreover, the interpretability of DBCF remains limited because real-world forgery patterns can be highly complex and may not be cleanly attributed to either localized manipulation traces or global semantic inconsistencies. Consequently, although the two branches are designed to model complementary representations, the current framework does not explicitly identify how different cues interact in each detection decision. Future work will investigate efficient backbone replacement, model compression or distillation, and cue visualization or forgery localization mechanisms to improve deployment efficiency and provide more explicit evidence for detection decisions.\\

\section*{CRediT authorship contribution statement}
\textbf{Fengming Gu:} Formal analysis, Investigation, Methodology, Project administration, Software, Validation, Writing – original draft Writing, – review \& editing. \textbf{Mingjie He:} Funding acquisition, Supervision, Validation, Writing – review \& editing.  \textbf{Zonghui Guo:} Funding acquisition, Supervision, Writing – review \& editing. \textbf{Jie Zhang:} Funding acquisition, Project administration, Writing – original draft, Writing – review \& editing. \textbf{Shiguang Shan:} Supervision, Writing – review \& editing.

\section*{Acknowledgments}
This work is partially supported by the Strategic Priority Research Program of the Chinese Academy of Sciences (No. XDB0680202), the Beijing Nova Program (No. 20230484368), the National Natural Science Foundation of China (No. 62276249), the National Natural Science Foundation of China (No. 62306298), the TaiShan Scholars Youth Expert Program of Shandong Province (No. tsqn202507108), and the Youth Innovation Promotion Association of the Chinese Academy of Sciences.

\bibliographystyle{elsarticle-num}

\bibliography{ref}

\end{document}